\documentclass{article}

\usepackage{arxiv}

\usepackage[utf8]{inputenc} 
\usepackage[T1]{fontenc}    
\usepackage{hyperref}       
\usepackage{url}            
\usepackage{booktabs}       
\usepackage{amsfonts}       
\usepackage{nicefrac}       
\usepackage{microtype}      
\usepackage{lipsum}		
\usepackage{graphicx}
\usepackage{natbib}
\usepackage{doi}
\usepackage{subcaption}
\usepackage{amsmath}

\title{Multi-Task Learning Improves Performance in Deep Argument Mining Models}

\author{Amirhossein Farzam \\
  Duke University \\
  \texttt{a.farzam@duke.edu} \\\And
  Shashank Shekhar \\
  New York University \\
  \texttt{shashank.shekhar@nyu.edu} \\ \AND
  Isaac Mehlhaff \\
  The University of Chicago \\
  \texttt{imehlhaff@uchicago.edu} \\\And
  Marco Morucci \\
  New York University \\
  \texttt{marco.morucci@nyu.edu}}

\renewcommand{\shorttitle}{Multi-Task Learning Improves Performance in Deep Argument Mining Models}

\hypersetup{
pdftitle={Multi-Task Learning Improves Performance in Deep Argument Mining Models},
pdfsubject={Argument Mining, NLP, CS},
pdfauthor={Shashank Shekhar},
pdfkeywords={Argument Mining, Multi-Task Learning, Large Language Models},
}

\begin{document}
\maketitle

\begin{abstract}
	The successful analysis of argumentative techniques from user-generated text is central to many downstream tasks such as political and market analysis. Recent argument mining tools use state-of-the-art deep learning methods to extract and annotate argumentative techniques from various online text corpora, however each task is treated as separate and different bespoke models are fine-tuned for each dataset. We show that different argument mining tasks share common semantic and logical structure by implementing a multi-task approach to argument mining that achieves better performance than state-of-the-art methods for the same problems. Our model builds a shared representation of the input text that is common to all tasks and exploits similarities between tasks in order to further boost performance via parameter-sharing. Our results are important for argument mining as they show that different tasks share substantial similarities and suggest a holistic approach to the extraction of argumentative techniques from text.
\end{abstract}


\section{Introduction}
\label{sec:intro}

Text content generated by online users is a fundamental source of information for understanding the ideas, feelings, and behavior of large populations of interest for social scientists. Within these texts, it is important to be able to recognize ideas and worldviews expressed by individuals on a large scale. To this end, argument mining (AM) has emerged in recent years as a sub-field of natural language processing (NLP) focusing on creating language models capable of detecting and classifying argumentative strategies in online texts. 

Within AM, several different sub-tasks have been proposed through the years. For example, \citet{Misra2013} focus on identifying agreement and disagreement in online texts, \citet{oraby2017and} propose a method to distinguish factual from emotional argumentation techniques, \citet{lawrence2017harnessing} detect the presence of certain rhetorical figures in arguments, and \citet{wachsmuth2017argumentation, wachsmuth2017computational} produce measures of argument quality. These are only some examples of the many distinct sub-tasks that have been identified in AM. In this paper, we suggest that all these AM sub-tasks share substantial similarity and use this idea to formulate a model that achieves high accuracy in several of these problems. 

More specifically, existing work in AM treats many of the sub-tasks within the field as separate problems and focuses on fine-tuning bespoke models for each task \citep[e.g.][]{Abbott2011, Stab2014a, stab2017recognizing, sheng2020nice}. While this approach has been demonstrated to work in many settings, it fails to take advantage of the substantial overlap and similarities between AM sub-tasks. 

\begin{figure}[!htb]
	\centering
    \includegraphics[width=0.49\linewidth]{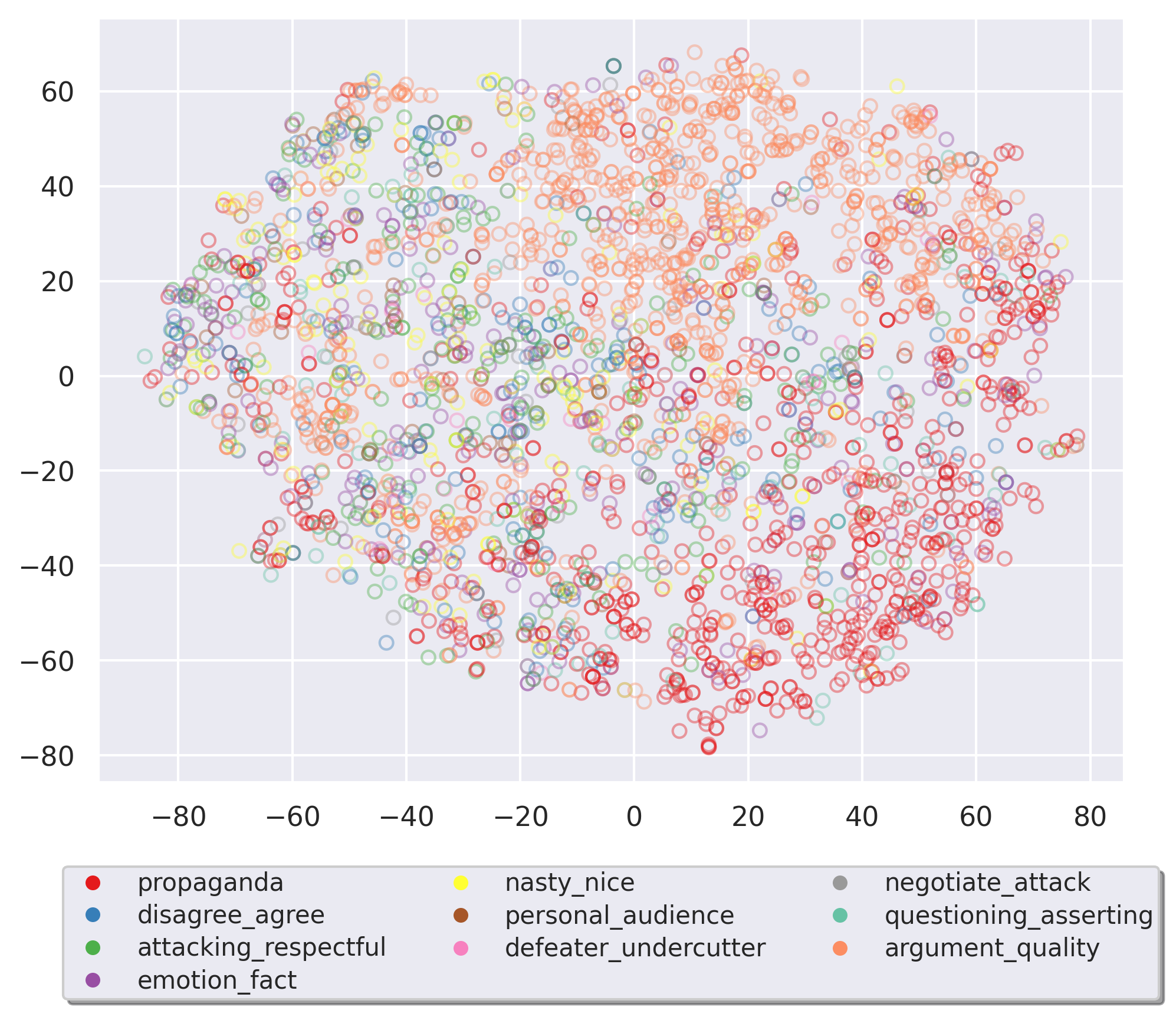}
	\caption{The t-SNE projection of the  BERT embedding included as the first layer in our model. Points are color-coded according to their task.}
	\label{fig:tsne_bert}
\end{figure}

In this paper we propose to take advantage of the similarities across AM tasks by constructing a multi-task model that constructs a shared latent representation of the inputs for each task, and uses this representation to make more accurate predictions for each individual task. Our models also provide evidence that AM sub-tasks do indeed share substantial conceptual overlap; the latent representations of different tasks output by our model depicted in Figure \ref{fig:tsne_bert} clearly depict clusters of individual tasks as well as substantial overlap between these clusters in representation space, indicating that the same latent features are informative for multiple tasks. 

The model we propose achieves State-Of-The-Art (SOTA) performance on all tasks for which we had information on previous SOTA metrics, and it also surpasses individual-task models fine-tuned on similar architectures for these tasks. In addition, our models allow for substantial computational gains over individual-task models as they permit training inference for many outputs at once, instead of having to train and evaluate an individual model for each desired task. 

Overall, our results have the important implication for AM as a field that further research and model-building should not only focus on taking advantage of the structure of the specific task of interest \citep[e.g.][]{jin2022logical}, but also on incorporating information from similar tasks into the model for better performance. 

Our paper will proceed as follows: first, we review relevant work (section \ref{sec:litreview}), second we introduce the different AM tasks and respective data sources that we incorporate in our model (section \ref{sec:data}), as well as the state of the art performance on these tasks. Then, we introduce our proposed model architecture, loss function and training regime (section \ref{sec:methodology}). Finally, we present our empirical results (section \ref{sec:results}) and discuss conclusions and future work (section \ref{sec:conclusion}).

\section{Related Work}
\label{sec:litreview}

We build on an active research agenda in argument mining (AM)---the automated extraction of argumentative structure, reasoning, and features from text \cite{Habernal2017}. \citet{Cabrio2018} identify two stages in AM: identifying arguments within documents and classifying those arguments on their characteristics, such as supporting, attacking, or background information. Our work is situated in the second stage, involving the identification of features or typologies of arguments.

Much computational work in AM has investigated argumentation in online interactions \cite{Abbott2011,Rosenthal2015,Swanson2015}, due in part to the vast amounts of available data and the ease of collecting it. But some scholars have used news articles to construct corpora of propaganda and fact-checking \cite[e.g.,][]{da2019fine,Rashkin2017}. Still others have leveraged monologues like persuasive essays or legal decisions \cite{Stab2014a,Walker2019}. We incorporate all three types of data into our models to further show that tasks with different generation processes and textual characteristics nevertheless exhibit common semantic structure.

There is evidence that many natural language tasks share a common core \cite{radford2019language}, and models trained on one task tend to also perform well on others. \citet{halder2020task} demonstrate that multi-task approaches benefit model performance in several natural language tasks such as topic detection and sentiment analysis. Multi-task approaches have been somewhat rare within AM---for example, \citet{cheng2020ape} uses task similarities to extract arguments and rebuttals from peer review---but these other tasks likely share enough common structure with ours that performance on them may also be improved via multi-task learning.

A prevalent model architecture for multi-task learning within computer vision involves segregating the network into shared and task-specific components. 
This conventional structure, termed as a ``shared trunk'' \cite{crawshaw2020multi}, typically comprises a universal feature extractor, constructed of convolutional layers that are employed by all tasks, and a distinct output branch for each task \cite{zhang2014facial, dai2016instance, zhao2018modulation, liu2019end}.
Further enhancements on this shared trunk template have been made by \cite{zhao2018modulation} and \cite{liu2019end}, who incorporated task-specific modules into the original framework. 

This template is not confined to computer vision but is also prevalent in multi-task learning models in NLP.
Traditional feed-forward architectures, using the shared trunk template in combination with task-specific modules, have been utilized for multi-task NLP by a variety of researchers \cite{collobert2008unified,collobert2011natural,liu2015representation,liu2016recurrent}. 
These architectures bear a structural resemblance to their counterparts in computer vision, featuring a shared, global feature extractor followed by task-specific output branches. However, in the context of NLP, the features in question are text representations.

\section{Data}
\label{sec:data}

We draw on three benchmark corpora to create a dataset with a diverse number of argument characteristics. We take 8 tasks from the Internet Argument Corpus (IAC), a collection of posts extracted from several online debate and discussion forums \cite{Abbott2016, Walker2012}. Each post is annotated on a variety of characteristics, such as whether the post expresses disagreement or uses an emotion- or fact-based argument, with a value in $[-5, 5]$ on each characteristic. Some researchers have dichotomized these data by removing observations around the midpoint \citet{Oraby2015}. This practice is not appropriate in the multi-task setting, however, as it would remove too much information that the model could use to build shared representations across tasks. Instead, we dichotomize the data by simply cutting on the scale midpoint. 

A wide array of studies have used the IAC to construct unique tasks \cite{Galitsky2018, Hartmann2019, Misra2016} and train single-task models \cite{Lukin2017, Misra2013, Oraby2016}. We are aware of comparable state-of-the-art benchmarks for three tasks: On the disagreement classification task, \citet{Abbott2011} achieve an accuracy of $68.2$\% and \citet{Wang2014a} achieve an F1 score of $69.6$\%. On the emotional or factual argument classification task, \citet{Oraby2015} achieve an F1 score of $53.7$\%. Finally, on the nasty or nice tone classification task, \citet{Lukin2013} achieve an F1 score of $69$\%. 

The second benchmark corpus we draw on is IBM-Rank-30k, a corpus of crowd-sourced arguments \citet{Gretz2020}. Two quality scoring functions then translated binary annotations into a continuous value of argument quality in $[0, 1]$. We use scores produced by the authors' weighted average scoring function because it accounts for coder reliability, leading to less noisy annotations. As with the IAC labels, we dichotomize the data by cutting on the scale midpoint.

The final corpus is introduced by \citet{da2019fine}, who collect articles from both propagandistic and non-propagandistic news sources and annotate sentences within each article that contain one or more of 18 different propaganda techniques, such as loaded language or causal oversimplification. We extract all sentences from each article, including those that are annotated as containing no propaganda techniques. Data from all three corpora are combined to create our final dataset. We use $80$\% for training and set aside $10$\% each for validation and test sets.

Finally, to help guard against overfitting, we conduct four types of data augmentation on the training set \cite{Shorten2021}. In back-translation, we translate the text into a different language, then translate it back to the original language. We choose German as the target language for its high lexical similarity to English. In contextual word embedding, we randomly choose thirty percent of tokens, feed the surrounding words to BERT, and substitute the predicted word in for the original. In synonym augmentation, we randomly choose thirty percent of tokens and substitute the most similar word from the WordNet lexical database \cite{Fellbaum1998}. Finally, in random cropping, we randomly delete thirty percent of tokens. Table \ref{tab:data} shows the total number of observations in the training set as well as the class balance for each task.

\begin{table}
\centering
\begin{tabular}{@{\extracolsep{0pt}}lcc}
\hline
\textbf{Task} & \textbf{Training N} & \textbf{Balance}\\
\hline
Propaganda & 61,909 & 63/37 \\ 
Disagree/Agree & 66,684 & 21/79 \\
Emotion/Fact & 76,403 & 41/59 \\
Attacking/Respectful & 65,998 & 66/34 \\
Nasty/Nice & 65,829 & 73/27 \\
Personal/Audience & 24,749 & 25/75 \\
Defeater/Undercutter & 24,357 & 38/62 \\
Negotiate/Attack & 26,604 & 44/56 \\
Questioning/Asserting & 29,791 & 66/34 \\
Argument Quality & 96,036 & 6/94 \\
\hline
\end{tabular}
\caption{Size and class balance of training data.}
\label{tab:data}
\end{table}

\section{Methodology}
\label{sec:methodology}

Our methodology is based on a multi-task learning approach which leverages the shared information across tasks corresponding to different sources of data, leading to improved performance on each task in a multi-label classification setting. 
The model architecture and the loss function are the two key components of our methodology.
Additionally, we make use of several standard training and optimization techniques, described in this section, in order to improve the performance.

\subsection{Model Architecture}

Our model architecture shares a key similarity to network templates comprising a shared trunk feeding task-specific modules, common to multitask learning architectures proposed in previous works \citep[e.g.][]{zhao2018modulation, liu2015representation}.
This architecture aims to utilize shared information across tasks through the shared trunk while learning distinct task features through the task-specific modules.
Following the same principle, we use a network with double-branching in layers following the shared trunk, in order to make use of commonalities across different types of tasks as well as more fine-grained information about each individual task.

\begin{figure}[!htb]
	\centering
	\includegraphics[width=.49\linewidth]{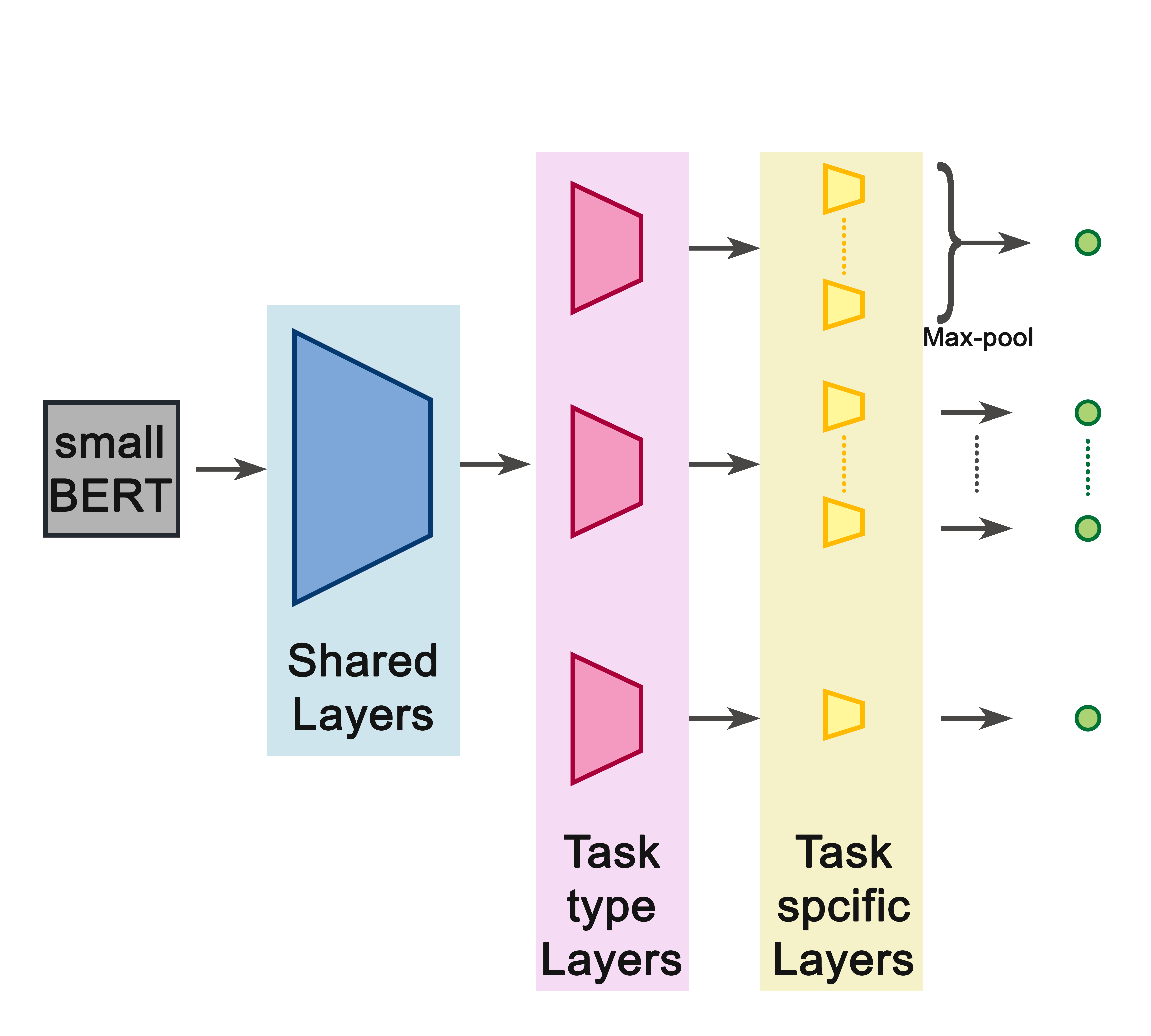}
	\caption{Model architecture.}
	\label{fig:architecture}
\end{figure}

We therefore use a feed-forward neural network with four sequential sets of layers: a base text embedding model shared across all tasks, followed by a shared encoder, which is followed by a double branching structure feeding two sets of task-specific modules. 
Our model architecture is based on the BERT model \cite{devlin2018bert}, which is a transformer-based model known for its effectiveness in various natural language processing tasks \cite{Kovaleva2019}. 
We use small BERT \cite{turc2019} as the base of our model, followed by three dense layers each followed by dropout. 
These layers help in learning features that are shared across tasks. 
The architecture then branches out to learn task-type and task-specific features. 
In particular, the architecture consists of four sets of layers, described below, and visualized in Figure \ref{fig:architecture}.
Each dense layer in the network uses a ReLU activation, with the exception of the final activation layer which is a sigmoid for binary classification.


\begin{itemize}
    \item \textbf{Shared embedding layers:}
    We use a BERT model \cite{devlin2018bert} to obtain an embedding of the text input.
    In order to keep the model size small and training practical, we use small BERT \cite{turc2019}, which outputs a 128-dimensional embedding.  
    The embedding model, shared across all tasks, is fine-tuned through our training.
    
    \item \textbf{Shared encoding layers:} 
    In addition to the base embedding model, all tasks share an encoder, consisting of two sequential dense layers each followed by a dropout layer.
    This helps learn a shared representation, used by all tasks, while allowing for sparsity and reducing the problem to learning our target features.
    
    \item \textbf{Task-type Layers:} 
    The first branching in the network architecture follows the shared layers aiming to learn coarse-grained task-specific features which are expected to share logical structures across tasks within each type.
    This is particularly suitable for multi-task learning on data consisting of a mixture of datasets, where the number of labels exceeds the number of sources in the mixture.
    Given such input data, in the first step towards learning the shared representation, the task-type layers learn dataset-specific features, while still utilizing commonalities between individual tasks sharing a dataset.
    For each task-type branch, we use two sequential dense layers each followed by dropout.
    Since we have three sets of target labels each corresponding to their own dataset, we use three main branches.
    
    \item \textbf{Task-specific Layers:} 
    Each main branch further branches out into individual task layers. 
    These layers aim to learn more fine-grained features from the representations produced through the main branches, and output a vector representation for each task.
    Each task-specific branch contains two sequential dense and dropout layers, which feed a sigmoid activation layer for predicting labels.
    The number of these sub-branches equals the number of individual features in the combined dataset. 
    In the branch corresponding to propaganda techniques, we additionally use a maximum pooling layer to reduce the 18 individual propaganda technique labels to a single binary propaganda classification, predicting whether a propaganda technique is used.
\end{itemize}

The full architecture is illustrated in Figure \ref{fig:architecture}.
Using this architecture, we obtain a vector representation of the size of the fine-grained features described in the dataset.
Note that this does not need to be the same as the size of the target output. It is not in this case, as we apply max-pooling to 18 entries of the output corresponding to propaganda techniques in order to obtain a binary label.
The network outputs a real-valued 10-dimensional vector which is then mapped to a binary vector of size 10 using individual thresholds for each label.
For the results produced in the main text of this paper we use a model with 17024 trainable parameters in addition to the parameters in small BERT.

\subsection{Loss Function}

The loss function plays a crucial role in our multi-task learning approach, which relies on a mixed corpora corresponding to different task types. 
The custom loss function is designed to handle the data size imbalance across task types, in addition to class imbalance, in order to effectively capture the contribution of each prediction to the overall loss. 
Considering this, given predicted labels $\hat{y}$ and true labels $y$, the total loss $\mathcal{L}$ used in our gradient descent optimization is:
\begin{align*}
    \mathcal{L}\left(\hat{y} | y \right)
    &=
    \sum_{k} \nu_{k} \mathcal{L}{\left( \hat{y} | y, \mathcal{D}_k\right)},
\end{align*}
where $D_{k}$ denotes the set of data point indices corresponding to task-type $k$, and $\nu_k \sim 1/|D_{k}|$ are the task-type weights.
The loss for each task type $k$, which accounts for the class imbalance across output labels, is:
\begin{align*}
    \mathcal{L}{\left( \hat{y} | y, \mathcal{D}_k\right)}
    &\sim
    \frac{1}{|T_k|}
    \sum_{j\in D_k} 
    \sum_{t\in T_k} 
    \sum_{c \in \mathcal{C}_t}
    w^c_t \  
    l{\left( \hat{y}_j | y_j = c \right)},
\end{align*}
where $l(.)$ is the loss function, $T_k$ denotes the set of tasks within task type $k$, and $\mathcal{C}_t$ is the corresponding set of classes. 
The class weights $w^c_{t}$, which are proportional to the inverse of the enrichment of class $c$ in task $t$ within dataset $k$, counter the impact of class imbalance.  
We use the binary cross-entropy loss for the loss function $l$ throughout our computations.
In the implementation, the loss computation is vectorized using masked matrices to filter entries by task.

\subsection{Model Training}

For training the parameters in our model, we take advantage of an array of optimization and training enhancement techniques.
We use an AdamW optimizer \cite{loshchilov2017decoupled} for the stochastic gradient descent with an initial learning rate of $0.0003$. To help avoid overfitting, we employ a weight decay rate of $0.01$ and 40\% dropout.
We use 5\% of data for warmup, a batch size of 256, and stop training after two epochs without a decrease in loss. 
We also incorporate threshold tuning, maximizing true positive rate while minimizing false positive rate, for optimal mapping of the sigmoid layer's output to binary labels.
All training hyperparameters are tuned through a standard grid search over $72$ sets of hyperparameters and selected based on F1 score in the validation set.

\section{Empirical Results}
\label{sec:results}

We evaluate our multi-task model's performance in terms of prediction metrics, computational efficiency, and comparison against state-of-the-art (SOTA) metrics on the target labels. 
We also offer evidence that the tasks we combine do indeed share important similarities by presenting
text embeddings and intermediate layer representations, in Figure \ref{fig:tsne_bert} and Figure \ref{fig:tsne_layers}. 
We show that our model outperforms the SOTA on several tasks (Table \ref{tab:SOTA}), while being substantially more computationally efficient than single-task counterparts (Figure \ref{fig:efficiency}).

\subsection{Commonalities Across Tasks}

Our model was trained on three different corpora, described in section \ref{sec:data}, which we argue possess important semantic similarities. To provide evidence of our 10 tasks existing within a common representation space, we present t-SNE projections \cite{van2008visualizing} of the input text embeddings corresponding to each label at three different locations within the neural network. Figure \ref{fig:tsne_bert}, discussed in section \ref{sec:intro}, shows the t-SNE projection from the output of the BERT model we use as our base encoder. Points are color-coded according to their task. If our text data carried mutually exclusive information applicable only to the particular task for which it was labeled, we would see distinct clusters of representations in Figure \ref{fig:tsne_bert}. 

There is some minor evidence of clustering, particularly with respect to the propaganda and argument quality tasks, but even those tasks have observations spanning the entire representation space, and they clearly mix with other task representations. This suggests that the fine-tuned BERT model is learning representations that reflect similar semantic and logical structures across tasks. We also highlight the fact that the clustering behavior within tasks observable in the figure shows that our model's embeddings are not completely discarding task-specific structure. Rather, our model learns task-specific representations, and those representations exist within a common space with other task-specific representations, thus further lending evidence to the theory behind our approach.

\begin{figure}[!htb]
	\centering
	\begin{subfigure}{\textwidth}
		\centering
		\includegraphics[width=0.49\linewidth]{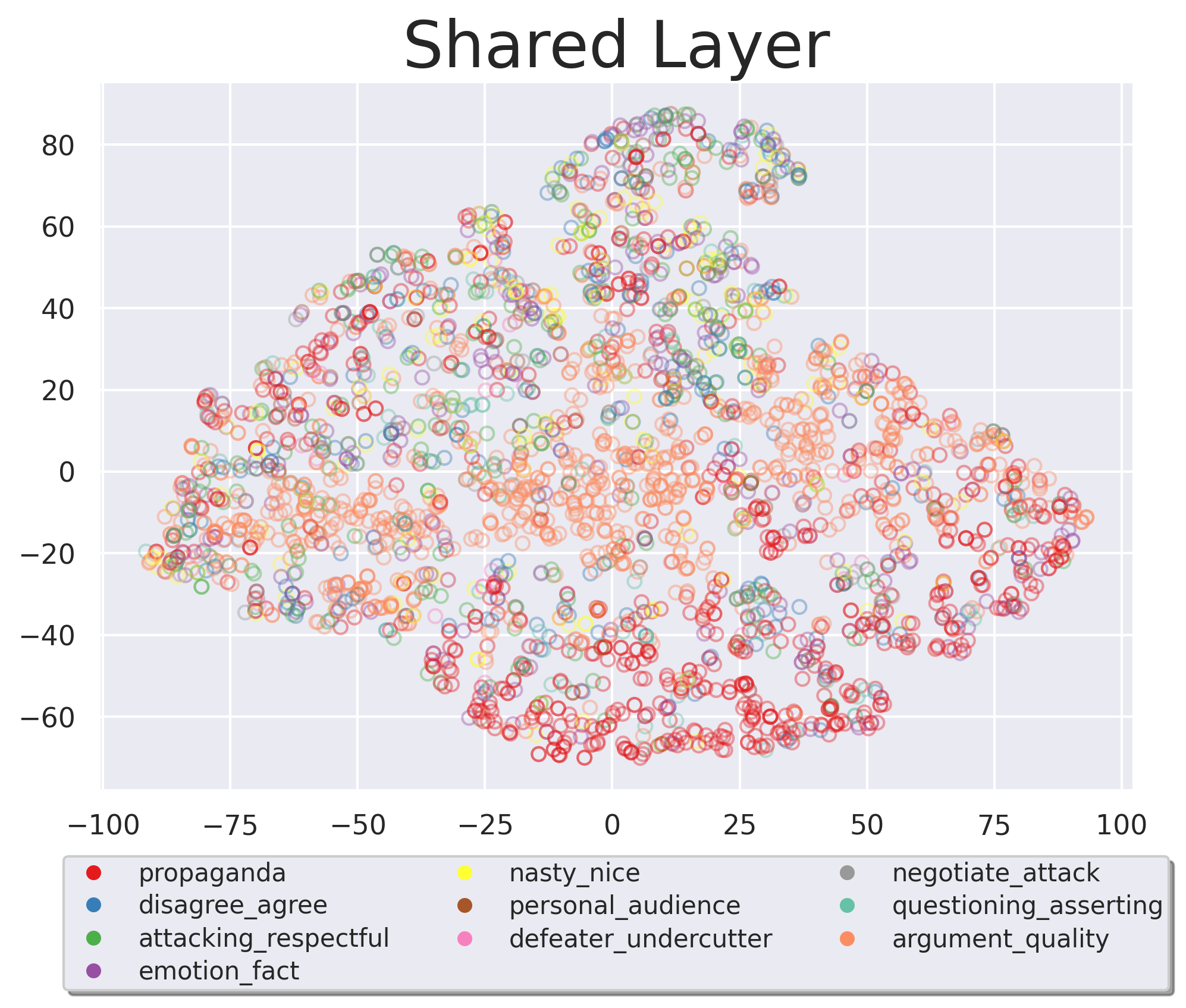}
		\caption{}
		\label{sfig:tsne_shared}
	\end{subfigure}%
	\\
	\begin{subfigure}{\textwidth}
		\centering
		\includegraphics[width=0.49\linewidth]{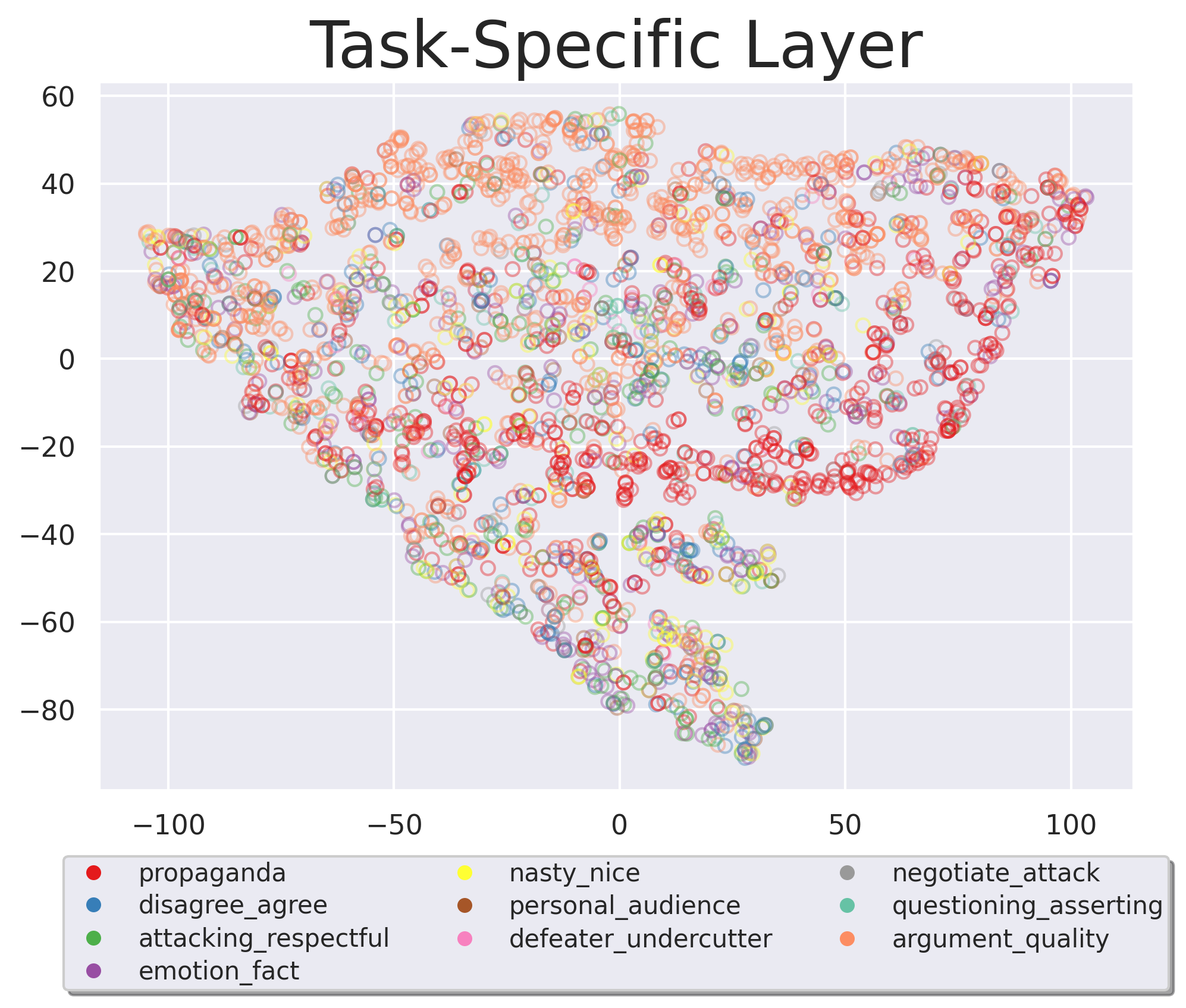}
		\caption{}
		\label{sfig:tsne_task}
	\end{subfigure}%
	\caption{ \footnotesize The t-SNE projection of representation obtained from the shared layers (top) and the task-specific layers (bottom).}
	\label{fig:tsne_layers}
\end{figure}

This pattern is largely preserved throughout the layers of our model. Figure \ref{fig:tsne_layers} presents similar t-SNE projections of two other intermediate layers: a shared layer (before any model branching occurs) and the final task-specific layer before the sigmoid activation (after the double-branching). Following the BERT model, each successive layer in the neural network gradually becomes more task-specific, and encodes information that is more relevant to distinguishing among tasks and among labels within tasks. It is notable, then, that we observe similar levels of clustering in the t-SNE projections regardless of model layer. Propaganda and argument quality tasks appear to inhabit more discernible regions of the representation space, but their clusters are neither well-defined nor tightly constrained. 

We take this consistent pattern as evidence that AM tasks share a common semantic space. Enabling a model to learn these fine-grained similarities and differences between tasks and across task types is therefore likely to improve performance relative to models that rely solely on shared features or no sharing at all. We test this conjecture in the next section.

\subsection{Performance Evaluation}

We evaluate the performance of our model primarily in terms of weighted F1 scores, which account for the class imbalances noted in Table \ref{tab:data}. In comparison with the SOTA for each task (Table \ref{tab:SOTA}), our model shows superior performance in predicting all of the tasks for which we had SOTA information available. This is a significant achievement, as it indicates that effectively leveraging shared features does improve prediction performance across multiple tasks. 

\begin{table*}
\centering
\resizebox{\textwidth}{!}{
\begin{tabular}{@{\extracolsep{0pt}}lcccccc} 
\hline \\[-1.8ex] 
& & & & & \textbf{Absolute} & \textbf{Relative} \\
\textbf{Task} & \textbf{Citation} & \textbf{Metric} & \textbf{Previous} & \textbf{New} & \textbf{Gain} & \textbf{Gain} \\
\hline \\[-1.8ex]
Propaganda & \citet{da2019fine} & F1 & 60.98 & 61.74 & 0.76 & 1.25 \\
Disagree/Agree & \citet{Wang2014a} & F1 & 63.57 & 71.38 & 7.81 & 12.29 \\
Disagree/Agree & \citet{Abbott2011} & Acc. & 68.20 & 70.73 & 2.53 & 3.71 \\
Emotion/Fact & \citet{Oraby2015} & F1 & 46.20 & 63.93 & 17.73 & 38.38 \\
Nasty/Nice & \citet{Lukin2013} & F1 & 69.00 & 73.69 & 4.69 & 6.80 \\  
\hline
\end{tabular}}
\caption{Comparison to previous state-of-the-art metrics.}
\label{tab:SOTA}
\end{table*}

Table \ref{tab:small_bert} shows a comparison of the predictive performance (as measured by the class-weighted F1-score) between baselines, single-task, and multi-task versions of the same model. The baseline metrics represent random guessing and the unigram metrics are produced by a naive Bayes classifier. While baselines underperform all the deep-network based approaches, the multi-task model also outperforms single-task models based on the same encoder architecture in all but three of the tasks that comprise our data. Again, we take this as evidence that our multi-task model is capable of exploiting the common structure between tasks in order to obtain more accurate predictions. In Table \ref{tab:comparison_base_model} in the Appendix, we show that this performance gain is not merely due to adding additional trainable parameters; multi-task models of various sizes perform comparably.

\begin{table*}
\centering
\begin{tabular}{@{\extracolsep{0pt}}lcccc}
\hline
\textbf{Task} & \textbf{Baseline} & \textbf{Unigrams} & \textbf{Single-Task} & \textbf{Multi-Task} \\
\hline
Propaganda & 55.47 & 38.46 & \textbf{63.07} & 61.74 \\ 
Disagree/Agree & 47.29 & 7.49 & 71.15 & \textbf{71.38} \\
Emotion/Fact & 45.80 & 21.91 & \textbf{68.11} & 63.93 \\
Attacking/Respectful & 56.47 & 51.16 & 67.46 & \textbf{68.07} \\
Nasty/Nice & 59.35 & 61.03 & 66.90 & \textbf{73.69} \\
Personal/Audience & 39.90 & 9.23 & 63.25 & \textbf{65.69} \\
Defeater/Undercutter & 53.4 & 45.21 & 45.97 & \textbf{55.65} \\
Negotiate/Attack & 36.93 & 55.31 & 64.76 & \textbf{64.81} \\
Questioning/Asserting & 50.57 & 57.47 & 59.61 & \textbf{63.23} \\
Argument Quality & 76.54 & 0.76 & \textbf{80.93} & 79.17 \\
\hline
\end{tabular}
\caption{Weighted F1 scores (with small BERT as base encoder). Baseline metrics are produced by random guessing and unigram metrics by a naive Bayes classifier.}
\label{tab:small_bert}
\end{table*}

We further investigated the impact of changing the base encoding model from small BERT to small ELECTRA \cite{clark2020electra} and base ALBERT \cite{lan2019albert}. Table \ref{tab:f1s} shows a comparison of performance across three different variants of our multi-task model. All three models have the architecture described in Section \ref{sec:methodology}, however, the base encoder differs each time. Generally, multi-task models trained on different encoders seem to display similar performance, indicating that the gain in performance due to the adoption of our framework is not necessarily due to the specific architecture of the encoder chosen. This is further demonstrated by the comparison of performance for different base encoders across individual tasks, which is offered in Table \ref{tab:compare_size} of the Appendix.  

\begin{table}
\centering
\begin{tabular}{@{\extracolsep{0pt}}lccc}
\hline
\textbf{Model} & \textbf{Prec.} & \textbf{Rec.} & \textbf{F1} \\
\hline
\multicolumn{4}{c}{Baselines}\\
\hline
Baseline & 62.26 & 52.43 & 52.17 \\
Unigrams & 33.65 & 44.55 & 34.80 \\
\hline
\multicolumn{4}{c}{Multi-Task Models}\\
\hline
Small BERT & \textbf{69.37} & \textbf{65.76} & \textbf{66.73} \\
Small ELECTRA & 69.19 & 63.98 & 65.16 \\
Base ALBERT & 58.65 & 63.10 & 58.34 \\
\hline
\end{tabular}
\caption{Comparison of base encoders. Metrics are class-weighted and averaged across tasks.}
\label{tab:f1s}
\end{table}

\subsection{Computational Efficiency}

\begin{figure}[!htb]
	\centering
	\captionsetup{position=top}
	\begin{subfigure}{\textwidth}
        \centering
        \includegraphics[width=0.49\linewidth]{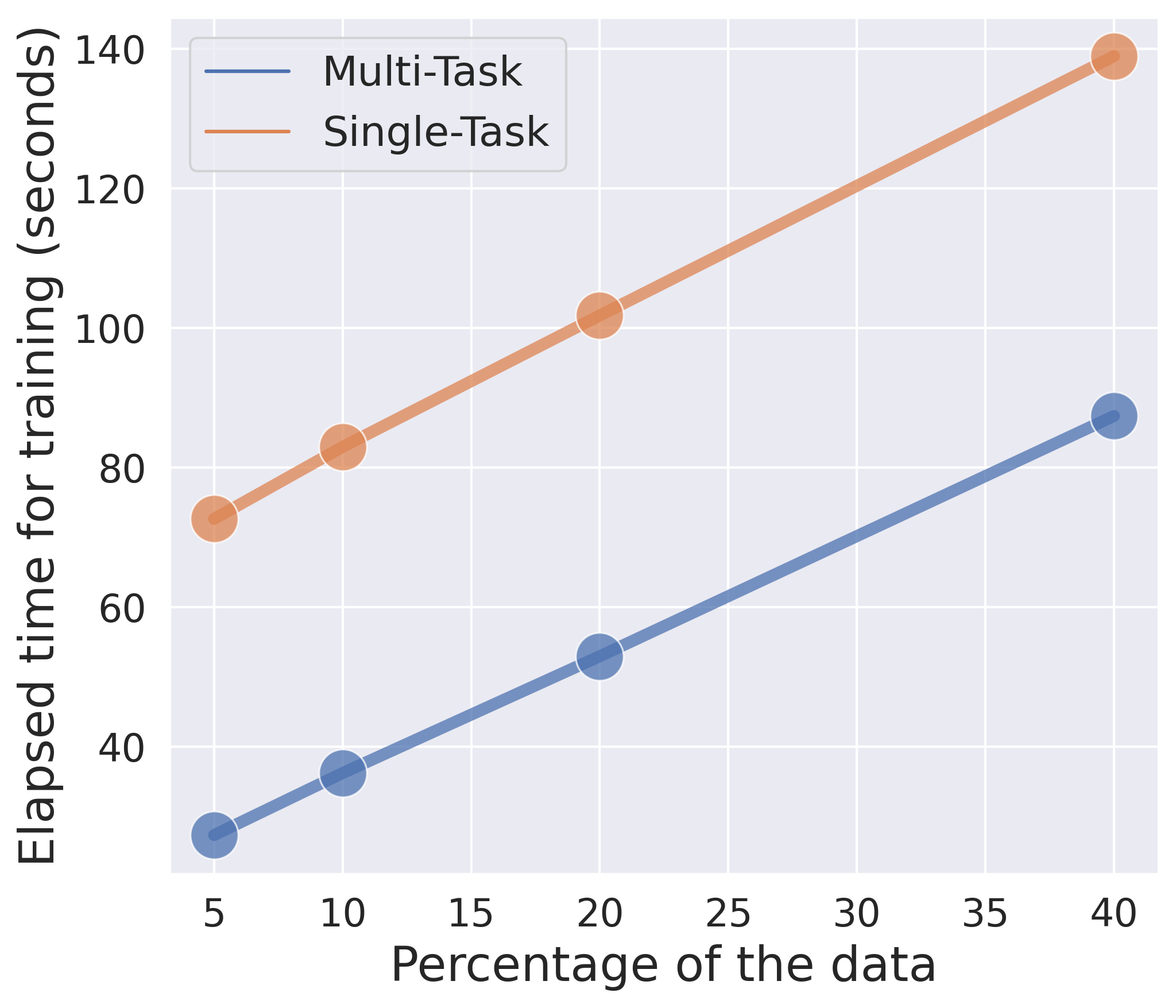}
		\caption{}
		\label{fig:efficiency_time}
	\end{subfigure}%
    \\
	\begin{subfigure}{\textwidth}
		\centering
		\includegraphics[width=0.49\linewidth]{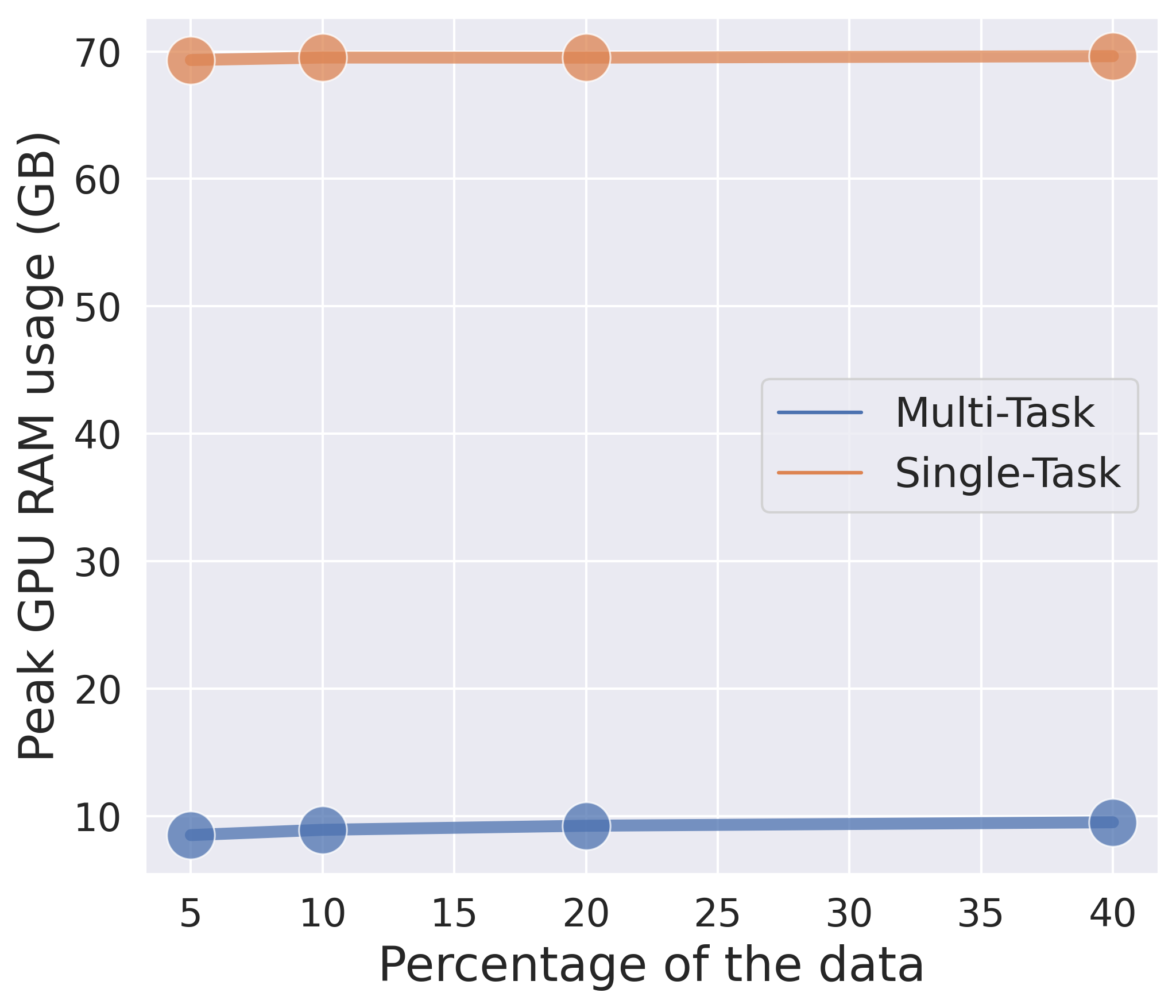}
		\caption{}
		\label{fig:efficiency_memory}
	\end{subfigure}%
	\caption{ \footnotesize Computational efficiency analysis. All models were run on one NVIDIA A100 GPU for one epoch.}
	\label{fig:efficiency}
\end{figure}

A key consideration, particularly when adding more trainable parameters as our model does, is whether the performance gain comes at the cost of more costly computation. We evaluate the peak GPU RAM usage and time to train our multi-task model and compare them to the same metrics from single-task models. We conduct this evaluation by randomly sampling $5$\%, $10$\%, $20$\%, and $40$\% of the training data to assess how computational load increases with data size. All models for this analysis were trained on one NVIDIA A100 GPU for one epoch. Figure \ref{fig:efficiency} displays the results. 

Our multi-task model achieves better performance using substantially lower computational resources, proving the branched task-specific modules in our model architecture to be an effective, yet practical, strategy for learning fine-grained features for label prediction.
Comparing our model's performance with single-task classification on individual tasks (Table \ref{tab:small_bert}), we observe that our model can achieve comparable performance while decreasing the computation time by at least $31$\%.

Put together, these observations indicate that this multi-task learning approach simultaneously has a performance and computational efficiency advantage over single-task models. Computational efficiency plots for different multi-task model sizes are included in the Appendix for comparison.

\section{Conclusion}\label{sec:conclusion}
Natural language tasks share substantial semantic and structural similarities, and deep learning models have been shown to be able to take advantage of these similarities in order to achieve better performance \cite{radford2019language}. In this paper, we have further extended this result to the field of argument mining. We have shown that AM tasks do indeed share a substantial amount of features, and that these shared features can be used to boost model performance across previously unrelated tasks. We have combined three data sources and proposed models that outperform the state-of-the-art on several of these tasks. Our models are also more computationally efficient and have overall better predictive accuracy than single-task models with comparable architectures. Aside from the practical usefulness of our models, our results are important for argument mining as a field, as they suggest that further research and model building should focus on exploiting commonalities between different tasks to boost performance. 

In future work, we propose to extend our analysis to several other AM tasks that share commonalities with those studied here \citep[e.g.][]{jin2022logical}, as well as devising improved model architectures for our multi-task setting. In particular, we propose to take advantage of frameworks such as contrastive learning \citep[e.g.][]{chen2020simple} to encode known similarities between tasks within the representations learned by the model. 

\section{Limitations}

As with all proposed models, ours carries important limitations. Although we show in the Appendix that the choice of base encoder does not have a drastic effect on performance, we suspect that the performance of our models is largely dependent on the ability to fine-tune a base encoder. Indeed, baseline models using unigram features performed quite poorly. Fine-tuning large base encoders---not to mention training one from scratch---can be computationally expensive. However, transfer learning may be able to help. Common semantic and logical structures across tasks point to opportunities for using transfer learning or pre-trained models from warm start to re-train on new tasks. 

Multi-task models also depend on data quality and sufficient semantic overlap across tasks. This is especially challenging in AM, as argument annotation is often highly subjective \citep[e.g.][]{Walker2012}, which can lead to noisy training data. Combining one low-quality dataset with other higher-quality ones may have a detrimental effect on model performance, as the model is unable to learn a shared representation space from noisy annotations, thus degrading performance on all tasks.

\bibliographystyle{unsrtnat}
\bibliography{references}  

\section{Additional Results}

In this appendix, we compare the performance of our multi-task model with alternative designs and configurations for multi-task learning, in terms of model architecture, network size, and the base encoder.

\subsection{Model Architecture}

\begin{figure}[!t]
	\centering
	\captionsetup{position=top}
	\begin{subfigure}{.45\textwidth}
		\captionsetup{position=top, labelformat=empty}
        \centering
        \includegraphics[width=\linewidth]{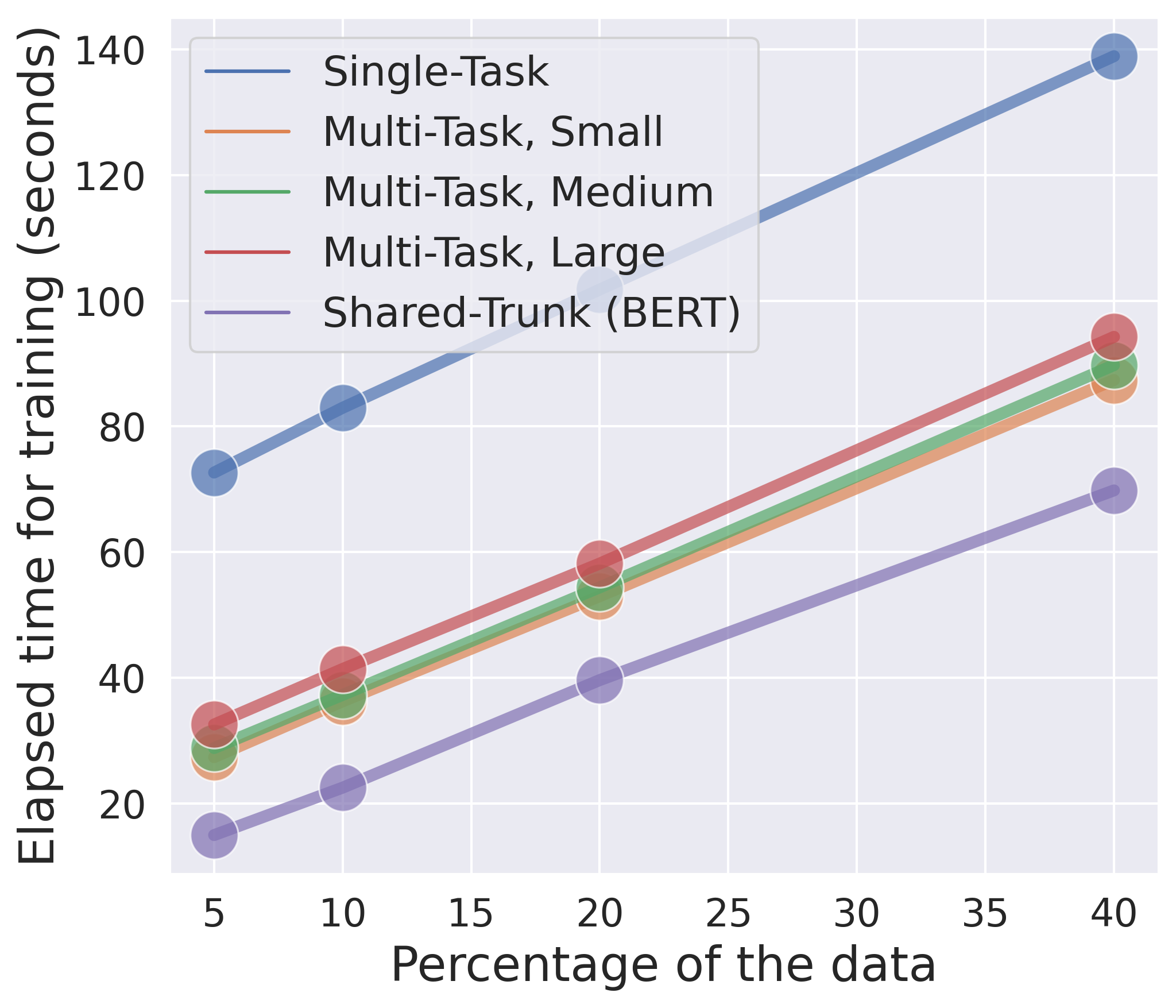}
		\caption{}
		\label{appendix_figfig:efficiency_time}
	\end{subfigure}%
    \\
	\begin{subfigure}{.45\textwidth}
        \captionsetup{position=top, labelformat=empty}
		\centering
		\includegraphics[width=\linewidth]{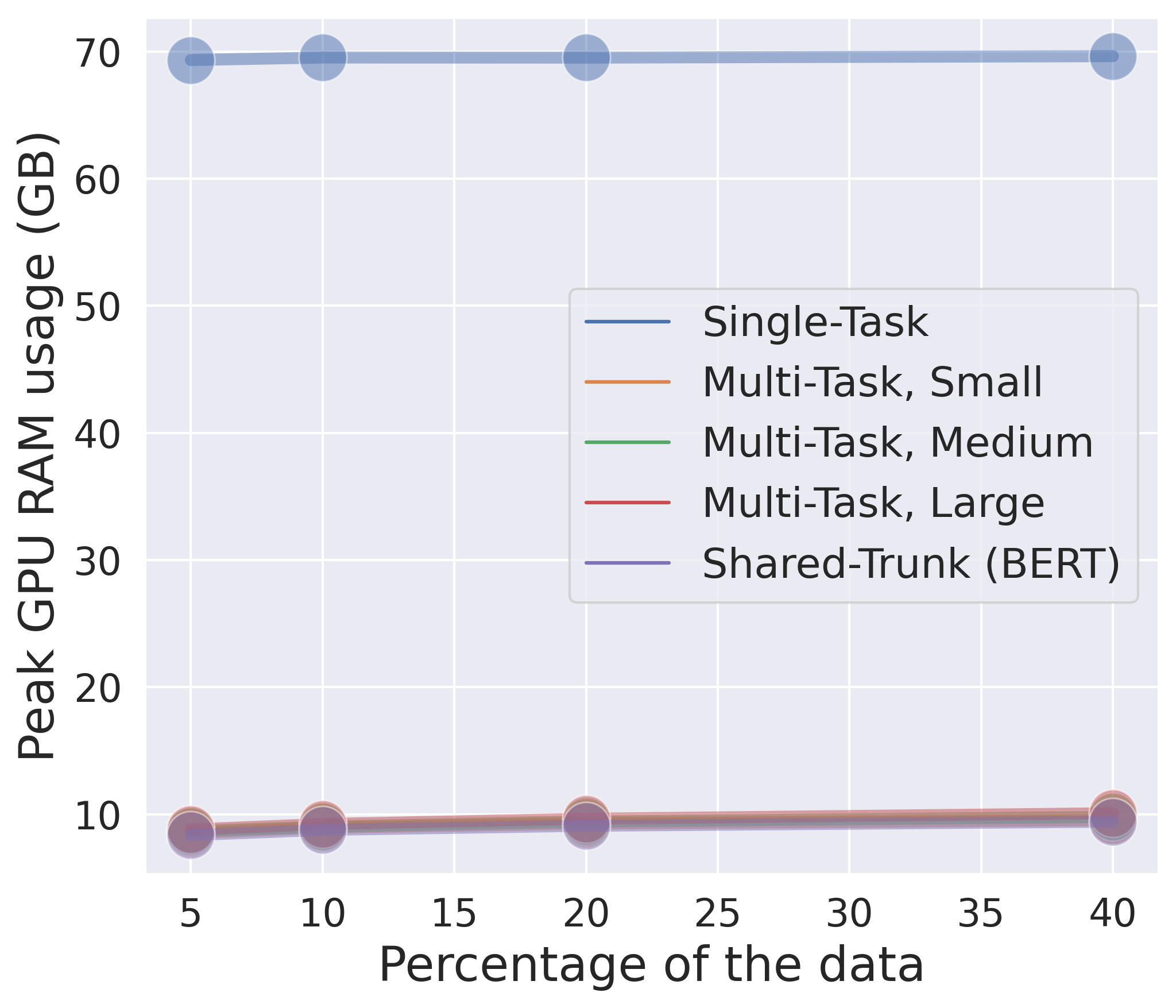}
		\caption{}
		\label{appendix_figfig:efficiency_memory}
	\end{subfigure}%
	\caption{ \footnotesize Computational efficiency for the single-task model as well as multi-task models with three different sizes of layers following the small BERT embedding.
    The small model contains 17024, medium 272384, and large 438784 trainable parameters in addition to the base encoder.}
	\label{appendix_fig:efficiency}
\end{figure}

Table \ref{tab:compare_size} compares the performance of our multi-task model---which incorporates branched task-type and task-specific modules---with a standard ``shared-trunk'' alternative, which consists of only a small BERT encoder followed by a sigmoid activation layer. This comparison shows the utility of our model architecture.
Our multi-task model outperforms the shared-trunk model on all but two tasks, where the F1 metric is within 1 percentage point of that of the shared-trunk model.
This performance gain comes at a negligible memory cost and a small increase in computation time (Figure \ref{appendix_fig:efficiency}).

\begin{table*}
\centering
\begin{tabular}{@{\extracolsep{0pt}}lcccc}
\hline
& \textbf{Shared Trunk} & \textbf{Multi-Task} & \textbf{Multi-Task} & \textbf{Multi-Task} \\
\textbf{Task} & \textbf{(BERT)} & \textbf{(17,024)} & \textbf{(272,384)} & \textbf{(438,784)} \\
\hline
Propaganda & 45.16 & 61.74 & 62.62 & 61.64 \\ 
Disagree/Agree & 37.96 & 71.38 & 62.07 & 66.74 \\
Emotion/Fact & 55.00 & 63.93 & 64.46 & 66.61 \\
Attacking/Respectful & 52.52 & 68.07 & 68.37 & 68.83 \\
Nasty/Nice & 55.62 & 73.69 & 73.04 & 73.38 \\
Personal/Audience & 66.51 & 65.69 & 70.17 & 65.24 \\
Defeater/Undercutter & 54.50 & 55.65 & 51.61 & 54.14 \\
Negotiate/Attack & 58.71 & 64.81 & 63.78 & 64.72 \\
Questioning/Asserting & 61.69 & 63.23 & 60.12 & 60.55 \\
Argument Quality & 79.34 & 79.17 & 68.36 & 81.28 \\
\hline
Average & 56.70 & 66.73 & 64.46 & 66.33 \\
\hline
\end{tabular}
\caption{Weighted F1 scores across shared layer sizes (with small BERT as base encoder). Number of trainable parameters in parentheses, not including base encoder.}
\label{tab:compare_size}
\end{table*}

\subsection{Network Size}

We also compare the performance of the small multi-task model we presented in the main text with alternative networks that preserve the same architectural design but increase the sizes of the layers, from $17024$ to $272384$ and $438784$ trainable parameters, following the base encoder. 
This comparison shows that the superiority in performance, due to the task-type and task-specific modules, is consistent across various network sizes and is not simply due to adding more trainable parameters on top of the shared trunk (Table \ref{tab:compare_size}).
Moreover, Figure \ref{appendix_fig:efficiency} further confirms that the layers following the BERT encoder are responsible only for a negligible increase in usage of computational resources, as multiplying the combined size of those layers by 16 (Multi-Task, Medium) and 32 (Multi-Task, Large) does not result in a substantial increase in elapsed time for training or peak GPU memory usage.

\subsection{Alternative Embedding Models}

In addition to comparing our model with other multi-task models, we also compare it to other base encoders. 
In particular, we deploy base ALBERT \cite{lan2019albert} and small ELECTRA \cite{clark2020electra}, replacing the small BERT encoder with each of these other base encoders in our multi-task model.
Although small BERT achieves the best average performance across different tasks, as the results in Table \ref{tab:comparison_base_model} suggest, using ELECTRA yields an average F1 score within 2 percentage points of that of small BERT, while ALBERT shows more variability across tasks with a lower average F1 score.

\begin{table}
\centering
\begin{tabular}{@{\extracolsep{0pt}}lccc}
\hline
& \textbf{Small} & \textbf{Small} & \textbf{Base} \\
\textbf{Task} & \textbf{BERT} & \textbf{ELECTRA} & \textbf{ALBERT} \\
\hline
Propaganda & 61.74 & 62.8 & 53.3 \\ 
Disagree/Agree & 71.38 & 59.4 & 69.2 \\
Emotion/Fact & 63.93 & 65.4 & 21.9 \\
Attacking/Respectful & 68.07 & 67.4 & 58.6 \\
Nasty/Nice & 73.69 & 71.5 & 61.1 \\
Personal/Audience & 65.69 & 68.5 & 64.1 \\
Defeater/Undercutter & 55.65 & 53.1 & 49.8 \\
Negotiate/Attack & 64.81 & 63.8 & 56.7 \\
Questioning/Asserting & 63.23 & 58.5 & 58.7 \\
Argument Quality & 79.17 & 81.2 & 90.0 \\
\hline
Average & 66.73 & 65.16 & 58.34 \\
\hline
\end{tabular}
\caption{Weighted multi-task F1 scores across base encoders.}
\label{tab:comparison_base_model}
\end{table}






\end{document}